\def\year{2020}\relax
\documentclass[letterpaper]{article} 
\usepackage{aaai20}  
\usepackage{times}  
\usepackage{helvet} 
\usepackage{courier}  
\usepackage[hyphens]{url}  
\usepackage{graphicx} 
\usepackage{graphicx}  
\usepackage{booktabs}       
\usepackage{amsfonts}       
\usepackage{amsmath}      
\usepackage{xcolor}

\title{Training Decision Trees as Replacement for Convolution Layers}
\author{\Large \textbf{Wolfgang Fuhl, Gjergji Kasneci, Wolfgang Rosenstiel, Enkeljda Kasneci}\\ 
	Eberhard Karls Universit\"at T\"ubingen\\
	Sand 14\\
	T\"ubingen, Germany\\
	wolfgang.fuhl;gjergji.kasneci;wolfgang.rosenstiel;enkelejda.kasneci@uni-tuebingen.de
}
 
\begin{document}

\maketitle

\begin{abstract}
	We present an alternative layer to convolution layers in convolutional neural networks (CNNs). Our approach reduces the complexity of convolutions by replacing it with binary decisions. Those binary decisions are used as indexes to conditional distributions where each weight represents a leaf in a decision tree. This means that only the indices to the weights need to be determined once, thus reducing the complexity of convolutions by the depth of the output tensor. Index computation is performed by simple binary decisions that require fewer cycles compared to conventionally used multiplications. In addition, we show how convolutions can be replaced by binary decisions. These binary decisions form indices in the conditional  distributions and we show how they are used to replace 2D weight matrices as well as 3D weight tensors. These new layers can be trained like convolution layers in CNNs based on the backpropagation algorithm, for which we provide a formalization.
	
	Our results on multiple publicly available data sets show that our approach performs similar to conventional neuronal networks. Beyond the formalized reduction of complexity and the improved qualitative performance, we show the runtime improvement empirically compared to convolution layers.
\end{abstract}

\section{Introduction and Related Work}
\noindent Conditioning CNNs is a modern approach to reducing runtime which is typically  achieved by activating only parts of the models or by pursuing the scalability of model complexity~\cite{ShazeerMMDLHD17,ioannou2016decision,chen2018neural} to reduce computational costs without compromising accuracy. Recent approaches even reduce the complexity of convolution layers \cite{keskin2018splinenets} without affecting the accuracy. This paper describes a new approach for the practical implementation of conditional neural networks using conditional distributions and binary decisions. Similar to \cite{keskin2018splinenets}, we replace convolutional layers to reduce computational complexity with the addition of indexing by simple binary decisions. We show analytically and empirically the reduction of the computational runtime on the basis of public data sets as well as the retention or increase of the accuracy of the model.

\begin{figure*}
	\centering
	\includegraphics[width=0.7\linewidth]{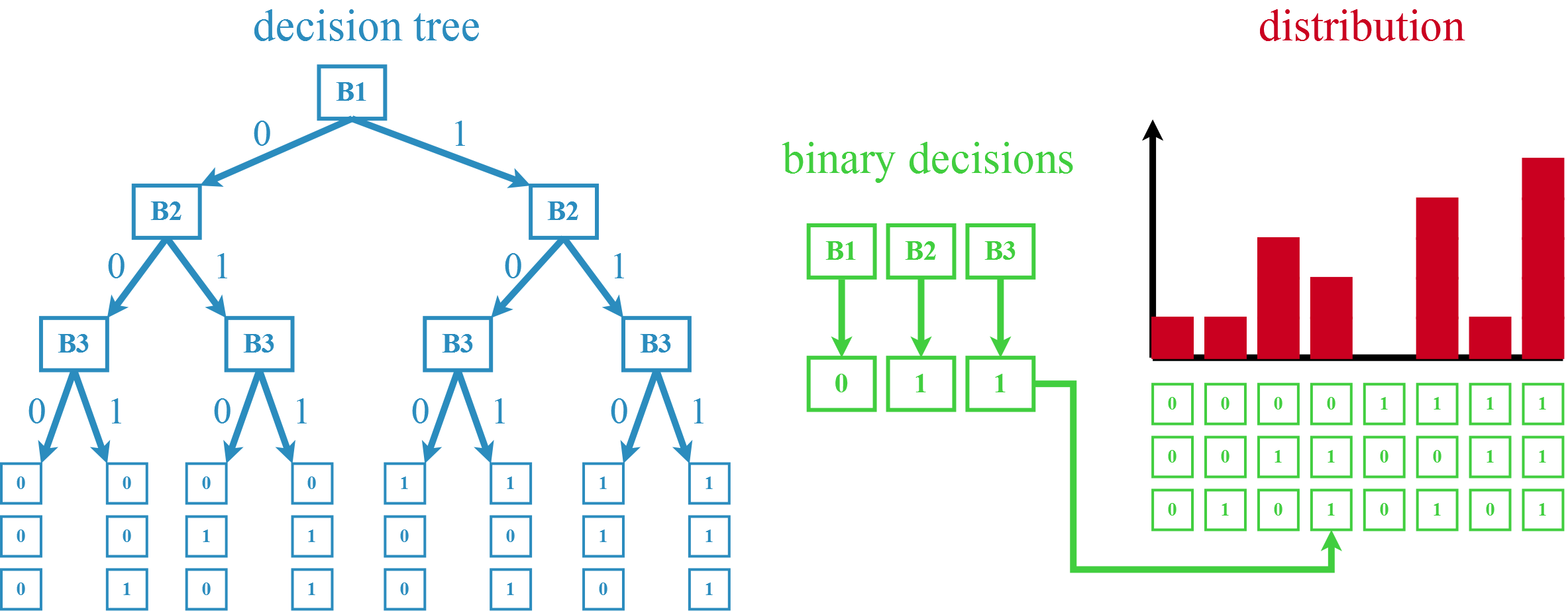}
	\caption{The left part depicts a decision tree. The decisions are true or false evaluations and the tree structure is complete and balanced. The middle part replaces the tree with a simple binary number and indexes a weight of a distribution. The distribution and its indices are shown on the right. Each value in the distribution corresponds to one leaf of the decision tree (left).}
	\label{fig:core}
\end{figure*}

There are four main categories of Conditional Neural Networks:
\begin{description}
	\item[1] Neural Networks that use loss functions for optimizing decision parameters.
	\item[2] Probabilistic approaches that learn a selection of experts.
	\item[3] Neural networks with decision tree architectures.
	\item[4] Replacement layers for the convolutions, which map hierarchical decision graphs conditionally to the input feature space.
\end{description}

The first category uses non-differentiable decision functions where the parameters for these are learned by an additional loss function. A loss function which maximizes the distances of the subcluster was presented in \cite{xiong2015conditional}. The path loss function is used in \cite{baek2017deep}. This is based on the purity of the data activation with respect to its class label distribution. The information gain is used in \cite{bicici2018conditional} to learn an evaluation function which allows to activate paths through the network.

In the second category, probabilistic approaches are pursued. Weights are assigned to each branch and treated as a sum over a loss function \cite{ioannou2016decision}. A similar approach is followed in \cite{ShazeerMMDLHD17}. The main difference is that a very high number of branches per layer is considered and the best k branches are followed in the training phase as well as in the test phase. Another approach trains two neural networks where one provides the decision probability at the output and the second network performs the classification \cite{kontschieder2015deep}. Both nets are trained jointly.

In the third category, the architecture of the neural network is similar to a decision tree. Randomized multi-layer perceptrons are used in \cite{rota2014neural} as branch accounts and trained together with the entire net. An alternative architecture is presented in \cite{denoyer2014deep}. Here, each account in a net has three possible subsequent nodes. The selection of the following node is done via an evaluation function which is learned via the REINFORCE algorithm \cite{denoyer2014deep}. In \cite{wang2017using}, partitioning features are learned which make it possible to train the whole network with the backpropagation algorithm. The architecture of the network corresponds to that of a binary decision tree. Each node in this network represents a splitting and has therefore exactly two outputs where only one can be active at a time \cite{wang2017using}.

The fourth and last category includes approaches that represent new layers in a neural network. Spatial transformation networks \cite{jaderberg2015spatial} learn a transformation of the input tensor, which simplifies further processing in the network. In general this is a uniform representation of the input tensor which can be understood as spatial alignment. Since the accuracy of a mesh depends not only on the input, but also on the filters in convolution layers, \cite{jia2016dynamic} introduces a layer that learns to generate optimal filters based on the input. This layer consists of a small neural network with convolution and transposed convolution layers. A further possibility for the conditional adaptation of neural networks is the configuration of the weights over a temporal course as it was realized in \cite{holden2017phase} over a phase function. The authors of \cite{holden2017phase} used a Catmull-Rom spline as phase function which can also be replaced by a neural network. The additive component analysis \cite{murdock2017additive} however tries to realize a non-linear dimension reduction by an approximation of additive functions. This is also defined as a fully connected layer and can be connected and trained in several layers. An approach based on this are the SplineNets \cite{keskin2018splinenets} which assign a new interpolated value to a learned spline via the response of a learned filter in the previous layer. This spline makes the function differentiable and several of these layers one behind the other can be understood as a topological graph.

Our novel approach is based on the idea of SplineNets \cite{keskin2018splinenets} to reduce convolutional complexity by simply mapping input characteristics to interpolated values. In addition, we simplify index generation with the general idea of binary neural networks \cite{courbariaux2015binaryconnect}. For this we use conditional distributions like random ferns \cite{bosch2007image}. The indices are determined based on simple larger, smaller comparisons between input values. These indices are used to select weights from several distributions and multiply them by the input values. The indices itself are the evaluation of the decision tree and the selected weight is the leaf node. This means that we consider both the values in the distributions and the input and output values as probabilities. This allows us to train the whole new layer with the backpropagation algorithm together with the whole net, as well as to connect several layers in series. The reduction of the computation complexity comes like with SplineNets \cite{keskin2018splinenets} by the indexing which has to be calculated only once and not like with convolution layers, where a new convolution has to be calculated for each filter. In addition, our layer does not have to learn function parameters or perform expensive multiplications to generate the indices.

Due to the conditional weights which are trained holistically in one layer, our approach belongs to category 4. Since the indices generation is based on comparisons and random ferns \cite{bosch2007image} represent a concretisation of random forests \cite{breiman2001random}, our approach also belongs to category 3. This means that it is a hybrid approach which is formalized as an independent layer but contains decision tree structures.

Our contributions in this work are:
\begin{description}
	\item[1] A new layer that selects leaf weights based on binary decisions.
	\item[2] The approximation of filters for index generation by binary decisions.
	\item[3] A differentiable formal definition of the forward execution which is suitable for the backpropagation algorithm.
	\item[4] Analytical and empirical evaluation of the quality and runtime improvement compared to CNNs.
\end{description}

\section{Methodology}
\begin{figure*}
	\centering
	\includegraphics[width=0.8\linewidth]{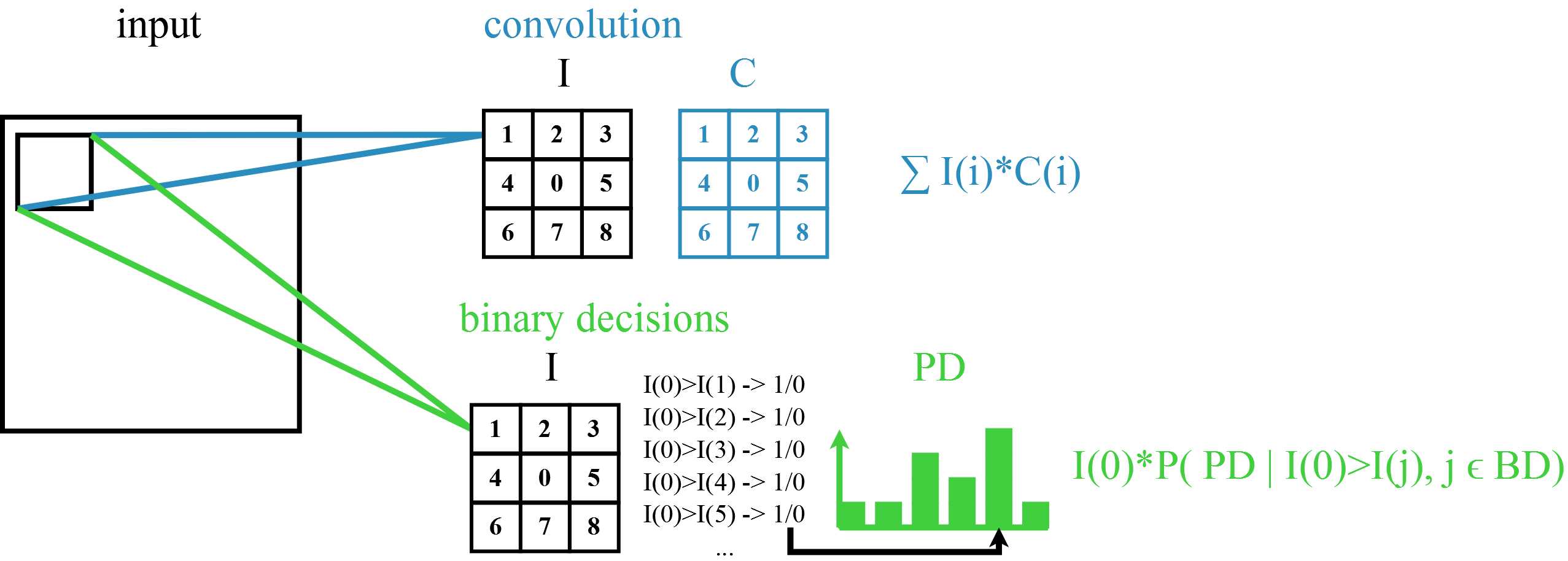}
	\caption{The functionality of the decision trees compared to a convolution. Like the convolution itself, an input window (I) is moved over the input tensor. In the upper part you can see the convolution which multiplies the weights (C) by the values in I and adds them up. In the lower part the decision tree is evaluated and the weight is multiplied by the central input value (0).}
	\label{fig:conv}
\end{figure*}

The Figure~\ref{fig:core} shows the core concept of our process. Random Ferns are binary decisions that are linked to conditional weights (see Figure~\ref{fig:core}). The binary decisions themselves represent the conditions. This means that it is a decision tree. Since each binary decision is always evaluated, the structure of this tree is arbitrary under the condition that each decision function must be contained once in each path, which makes the decision tree a balanced tree.

\begin{equation}
P(PD|I(k)>I(j), \forall j,k \in BD, j \ne k)
\label{eq:fernformula}
\end{equation}

Equation~\ref{eq:fernformula} describes the evaluation of such a decision tree or Fern. $I$ is the input tensor, $PD$ the distribution (see Figure~\ref{fig:core}) and $BD$ the indices of the comparisons. To use this decision tree now like a convolution the indices in $BD$ refer only to values in an input window which is moved over the whole input tensor (see Figure~\ref{fig:conv}). To combine several of these decision trees, the weights are multiplied. In the case of Equation~\ref{eq:fernformula}, this would be the centered input values at the current window position making it easy to determine the derivative and thus the gradient. Another simplification of Equation~\ref{eq:fernformula} is to compare all positions in $BD$ using only the central value (see Figure~\ref{fig:conv}). This simplifies the back propagation of the error.

\begin{equation}
I(0)*P(PD|I(0)>I(j), \forall j \in BD)
\label{eq:fernformulaconv}
\end{equation}

\begin{figure}
	\centering
	\includegraphics[width=0.99\linewidth]{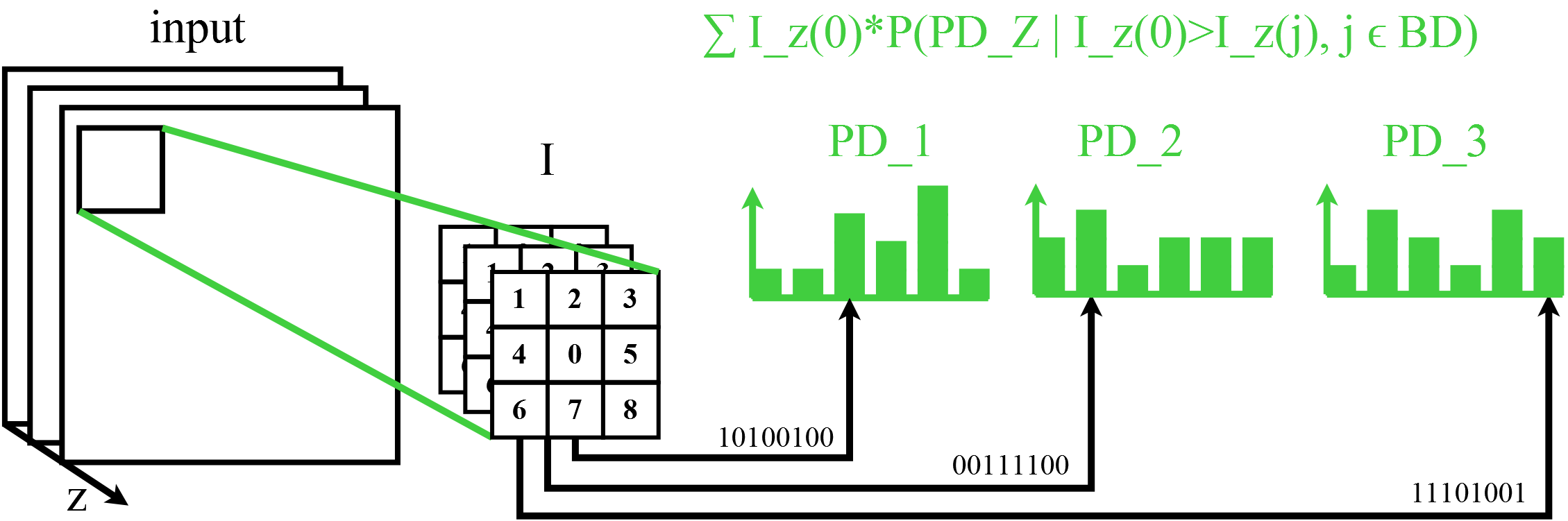}
	\caption{The use of multiple decision trees to process one input tensor. The indices of the comparisons ($BD$) are the same for all distributions ($PD$) which are different.}
	\label{fig:tensor}
\end{figure}

This leads to Equation~\ref{eq:fernformulaconv} which describes the evaluation of the decision tree for an input window. In the case of convolutions, this input window is not necessarily two dimensional, but also a tensor of weights. This tensor is represented by several distributions. Each depth value of the input tensor has its own distribution as with convolutions, where each depth uses its own two-dimensional weight matrix (see Figure~\ref{fig:tensor}).

This means that in the case of decision trees, each input depth has its own decision tree in the sense of its own distribution. For Equation~\ref{eq:fernformulaconv} this means that each depth of the input tensor $I_{i}$ with depth $z$ indexes its own distribution $PD_{i}$ over the same indexes $BD$.

\begin{equation}
\sum_{i=0}^{z-1}I_{i}(0)*P(PD_{i}|I_{i}(0)>I_{i}(j), \forall j \in BD)
\label{eq:fernformulatensor}
\end{equation}

Equation~\ref{eq:fernformulatensor} describes the calculation where it has to be taken into account that each depth performs a multiplication with the central value and at the end, as with convolutions, the sum of all multiplications is computed. This summation makes it easier to determine the gradients for each distribution because there are no multiplicative dependencies between the distributions.

\begin{figure*}
	\centering
	\includegraphics[width=0.8\linewidth]{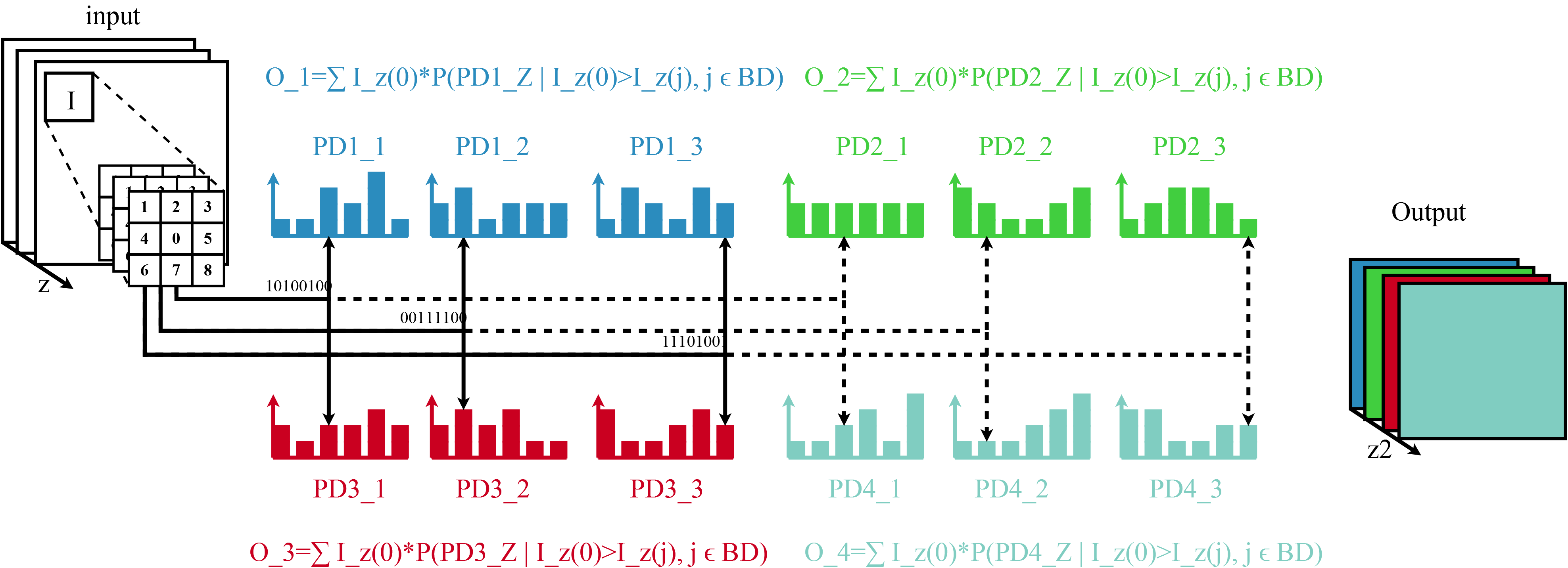}
	\caption{The use of different decision trees in multiple layers where the used indexes $BD$ are the same.  Each output layer is represented as its own color (red, blue, green, turquoise).}
	\label{fig:layerdepth}
\end{figure*}

The next step describes the layer depth of the decision trees so that these decision trees can now also be used like convolution layers in neural networks (see Figure~\ref{fig:layerdepth}). As in the previous step, the same indexes $BD$ are used for all layers but different distributions $PD_{l,i}$ are used for each layer. The reason for this is that the complexity of the calculation is reduced compared to convolutions.

\textit{Complexity: The calculation of a convolution layer with the input tensor t and n-times the convolution window c requires $t_x*t_y*t_z * n * c_x*c_y$ multiplications and additions. The decision trees, on the other hand, only have to determine the indices once, so that $n=1$ can be set, thus reducing the complexity by the output depths. In addition, the multiplications are replaced by simple larger or smaller comparisons and a multiplication. From this it follows that $t_x*t_y*t_z*|BD|$ comparisons are performed and $t_x*t_y*t_z*n$ multiplications and additions.}

\begin{equation}
O_l=\sum_{i=0}^{z-1}I_{i}(0)*P(PD_{l,i}|I_{i}(0)>I_{i}(j), \forall j \in BD)
\label{eq:fernformulalayerdepth}
\end{equation}

To extend Equation~\ref{eq:fernformulatensor} in this respect, each individual output layer $O_l$ must be assigned a set of distributions $PD_{l,i}$. Equation~\ref{eq:fernformulalayerdepth} describes this change, but it is important to make sure that every tree uses the same $BD$ indexes.

\begin{figure}
	\centering
	\includegraphics[width=0.99\linewidth]{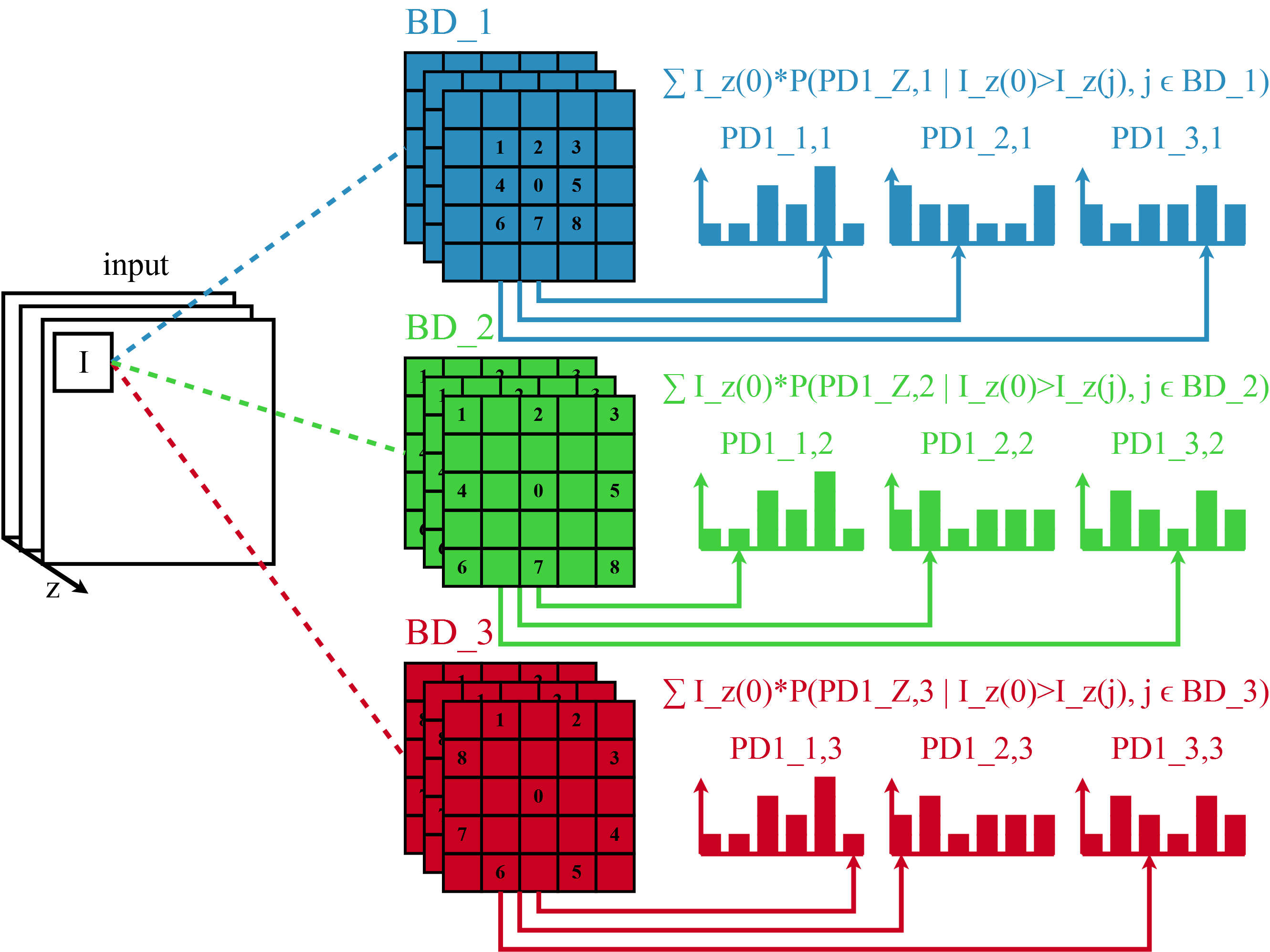}
	\caption{Using the inception architecture for training the decision trees. The indexes $BD_k$ are different for each color (red, green, blue) and are processed in parallel. Each of these index sets uses a distribution for each depth of the input tensor, as shown in Figure~\ref{fig:tensor}. The three index sets shown correspond to the output of one depth of the output layer. In the case of multiple output layers, the inception architecture is used as shown in Figure~\ref{fig:layerdepth}.}
	\label{fig:inception}
\end{figure}

A disadvantage of the approach presented so far is that the size of the distributions grows exponentially $2^{|BD|}$. This means that the memory requirements can very quickly reach the limits of modern computers and the numerical calculation of very small numbers in large distributions can become too inaccurate. Another disadvantage of large distributions, i.e. a large number of binary comparisons, is that the probability that an index will be used during training decreases the larger the distribution is. For a convolution of the size $5 \times 5$ a distribution size of $2^{24}$ would be needed, which contains all comparisons with the central value. In order to make it possible to use several small distributions and still make it possible to cover larger input windows, we use the idea of inception architecture~\cite{szegedy2015going}. This means that different index sets $BD_k$ with depth $b$ associated with different distributions $PD_{l,i,k}$ are aggregated in an output tensor. In our implementation we used the summation per layer.

\begin{equation}
\begin{split}
O_l=\sum_{k=0}^{b-1}\sum_{i=0}^{z-1}I_{i}(0)*P(PD_{l,i,k}|I_{i}(0)>I_{i}(j), \\
\forall j \in BD_k)
\label{eq:inception}
\end{split}
\end{equation}

Equation~\ref{eq:inception} describes the complete forward propagation per output layer of the presented new method for training decision trees in neural networks. All binary decision sets $BD$, with amount of sets $b$ are used to compute the index for the assigned distributions $PD_{ l,i,k}$. The sum of all selected weights in $PD_{ l,i,k}$ multiplied with their corresponding input value $I_{i}(0)$ is calculated for each input window and written into the output tensor $O_l$. The bias term itself is omitted in the formulas to simplify them but is used as in conventional convolution layers.

The backward propagation of the error occurs inversely to the forward propagation. This means that as with convolution layers, a convolution with the error tensor takes place for each input value of the input tensor.

\begin{equation}
\begin{split}
ERRI_{i}=\sum_{k=0}^{b-1}\sum_{l=0}^{L-1} ERRO_{l}(i)*P(PD_{l,i,k}| \\
I_{i}(0)>I_{i}(j), \forall j \in BD_k)
\label{eq:backpropagation}
\end{split}
\end{equation}

Equation~\ref{eq:backpropagation} describes the back propagation where $L$ is the depth of the output tensor. Thus each value of the input layer $i$ is assigned the sum of the errors $ERRO_{l}(i)$ multiplied by the indexed weights $PD_{l,i,k}$. In addition, for each value participated in the binary decisions the error is added divided by the size of the used binary decision set $BD_k$ (Equation~\ref{eq:backpropagation2}).

\begin{equation}
\begin{split}
ERRI_{i}(j,k)+=\sum_{l=0}^{L-1} ERRO_{l}(i)*P(PD_{l,i,k}| \\
I_{i}(0)>I_{i}(j), \forall j \in BD_k)/|BD_k|
\label{eq:backpropagation2}
\end{split}
\end{equation}

Equation~\ref{eq:backpropagation2} is calculated for each index $j$ in each used binary decision set $BD_k$ and sums the error over the output tensor of the depth $L$. The division by the record size $|BD_k|$ results in an equal share of the error being assigned to each index. This is due to the fact that the participation in the resulting error is independent of the binary value of the evaluation from the decision function.

\begin{equation}
Grad(PD_{l,i,k})=\sum_{l=0}^{L-1} I_{i}(0)*ERRO_{l}(i)
\label{eq:gradient}
\end{equation}

To determine the gradient, only the derivation between the generated error and the input needs to be considered. This is described in Equation~\ref{eq:gradient} and shows that only the central value of the input window and the output value are required. For the binary decision functions, the derivation is 0, since these are independent of the weights in the distribution.

\section{Experiments}
\begin{figure}
	\centering
	\includegraphics[width=0.6\linewidth]{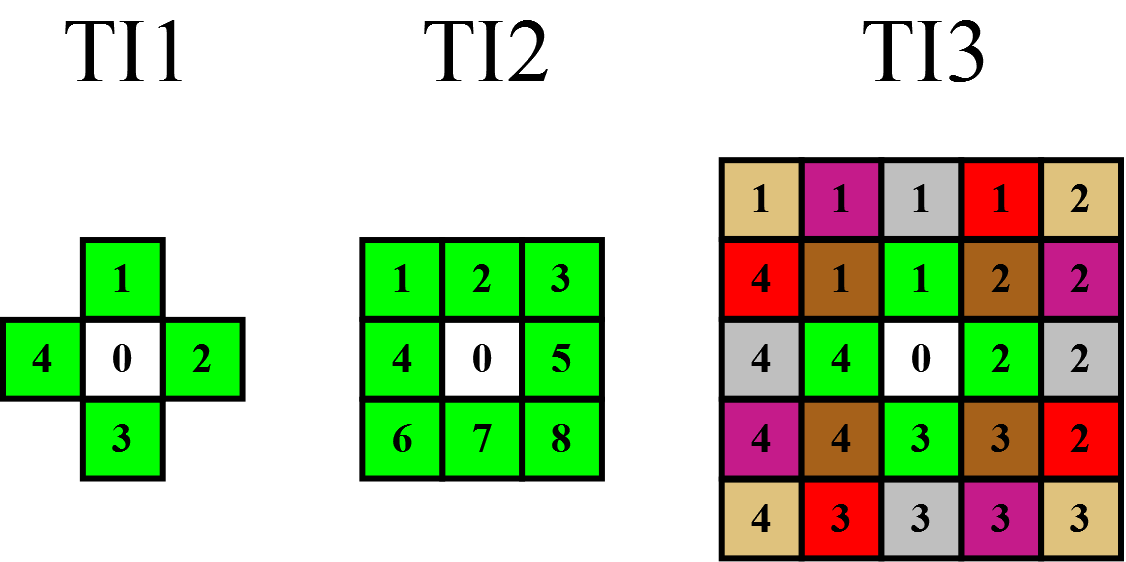}
	\caption{All index patterns ($BD$) used in our experiments. TI1 consists of four comparisons and the resulting distribution size of $2^4$. TI2 is the full approximation of a $3 \times 3$ convolution with distribution size $2^8$. TI3 is a pattern which uses the inception technique and consists of six distributions. Each pattern in TI3 has the same color and consists therefore out of four indexes. The full parameter size for one TI3 pattern is $6*(2^4)$.}
	\label{fig:arch}
\end{figure}

Figure~\ref{fig:arch} shows the used index patterns for our evaluations. We used the models LeNet-5~\cite{lecun1998gradient} with rectifier linear units (ReLu) instead of the hyperbolic tangent function and the deep residual model with depth 16 (ResNet-16) and 34 (ResNet-34). In both models (ResNet-16 and ResNet-34) we used a batch normalization block after each convolution or decision tree layer. The LeNet-5 model was used in the comparison on the MNIST~\cite{lecun1998gradient} dataset with the index patterns TI2 and TI3 (Figure~\ref{fig:arch}). The ResNet-34 was used for the comparison on the CIFAR10~\cite{krizhevsky2009learning} dataset with the TI1 pattern. As an alternative evaluation for image classification we used landmark regression. Therefore, we compared the decision trees with convolutions on the 300W~\cite{zhu2012face} dataset using the ResNet-16 and the TI1 patterns.

The general idea behind our experiments is not to surpass the state-of-the-art, but to compare decision trees with convolutions in the same architecture. For this purpose we tried to get as close as possible to the results of the state-of-the-art with simple means and to design the training process for convolutions and decision trees in the same way.

\paragraph{MNIST }
consists of 70,000 hand written digits and has therefore ten classes. Each image has a resolution of $28 \times 28$ pixels and is provided as gray scale image. The training set contains 60,000 and the test set 10,000 images. This data set is size-normalized and centered and represents a subset of the larger NIST dataset. As evaluation metric the classification accuracy ($\frac{Correct~Classification}{Test~set~size}$) is used. We only report the best result as it is done for the state-of-the-art~\cite{wan2013regularization,cirecsan2012multi,sato2015apac}.

\paragraph{CIFAR10 }
consists of 60,000 color images with ten different categories. Each image has a resolution of $32 \times 32$ pixels and is provided in the RGB format. The training set contains 50,000 and the test set 10,000 images. As evaluation metric the classification accuracy ($\frac{Correct~Classification}{Test~set~size}$) is used. We only report the best result as it is done for the state-of-the-art~\cite{DBLP:journals/corr/Graham14a,springenberg2014striving,mishkin2015all}.

\paragraph{300W }
is an aggregation of multiple datasets (LFPW~\cite{belhumeur2013localizing}, HELEN~\cite{le2012interactive}, AFW~\cite{zhu2012face} and XM2VTS~\cite{messer1999xm2vtsdb}).  The training set consists of 3,148 face images from the LFPW and HELEN dataset. For the test set 689 images are provided. Each image has 68 annotated landmarks~\cite{sagonas2013semi}. In the evaluation the test set is separated into three categories the full set, the challenging set (iBUG, 135 images) and the common set (LFPW and HELEN, 554 images). As evaluation metric we used the normalized mean errors (NME) which corresponds to the average distance between detected and annotated landmark, normalized by the pixel distance between both eye centers. This is the same evaluation procedure as the state-of-the-art~\cite{wingloss2018cvpr,dong2018style,ren2016face}.

\paragraph{Training parameters for MNIST: }
We used the Adam optimizer~\cite{kingma2014adam} with the first momentum set to 0.9 and the second momentum set to 0.999. Weight decay was set to $5*10^{-4}$ for the convolutions and to $10^{-8}$ for the decision trees. The batch size was set to 400 and each batch was always balanced in terms of available classes. This means that in each batch each class was represented 40 times. The initial learning rate was set to $10^{-2}$ and reduced by $10^{-1}$ after each 100 epochs until it reached $10^{-4}$. For the learning rate of $10^{-4}$ we continued the training for additional 1000 epochs and selected the best result. For data augmentation we used random noise in the range of 0-30\% of the image resolution.

\paragraph{Training parameters for CIFAR10: }
We used the Adam optimizer~\cite{kingma2014adam} with the first momentum set to 0.9 and the second momentum set to 0.999. Weight decay was set to $5*10^{-5}$ for the convolutions and to $10^{-10}$ for the decision trees. The batch size was set to 50 with the same batch balancing approach as for the MNIST dataset. For CIFAR this means each batch consisted of five examples per class. The initial learning rate was set to $10^{-2}$ and reduced by $10^{-1}$ after each 500 epochs until it reached $10^{-5}$. For the learning rate of $10^{-5}$ we continued the training for additional 1000 epochs and selected the best result. For data augmentation we used random cropping of $24 \times 24$ patches, random color offsets, random color distortion, flipping the image horizontally and vertically as well as random noise in the range of 0-20\% of the image resolution. Additionally, we overlayed patches of the same class with an intensity of up to 20\%.

\paragraph{Training parameters for 300W: }
All images where resized to $80 \times 80$ pixels. We used the Adam optimizer~\cite{kingma2014adam} with the first momentum set to 0.9 and the second momentum set to 0.999. Weight decay was set to $5*10^{-4}$ for the convolutions and to $10^{-8}$ for the decision trees. The batch size was set to 30. All 20 iterations an evaluation of the landmark accuracy was performed for the test and the training set. The accuracy on the training set was used to balance the batches. This was done by splitting the training set into three categories. The first category are the most inaccurate 20\%. For the second category we used the range between the first category and the most inaccurate 50\%. The last category is the range between the second category and the most inaccurate 80\%. For each batch we selected ten examples out of each category. The initial learning rate was set to $10^{-6}$ and increased by $10^{1}$ after each 100 epochs until it reached $10^{-4}$. For the learning rate of $10^{-4}$ we continued the training for additional 1000 epochs. Afterwards, we reduced the learning rate after each 100 epochs by $10^{-1}$ until we reached $10^{-7}$ and stopped the training. For data augmentation we used random noise in the range of 0-20\% of the image resolution. The image and landmarks where randomly shifted by up to 20\% of the image resolution into each direction. Additionally, we added randomly Gaussian blur ($\sigma=[1.0, 1.3]$). For occlusions we overlayed up to three boxes and filled them either with a fixed random value or a random value for each pixel in the box. We also randomly changed the contrast of the image in the range [-40, 40].

\paragraph{Hardware and implementation: }For training and evaluation we used two different hardware setups. For LeNet-5 we used a desktop PC with an Intel i5-4570 CPU (3.2 GHz), 16 GB DDR4 RAM, NVIDIA GTX 1050Ti GPU with 4GB RAM and Windows 7 64 bit operating system. The second hardware setup was used for the ResNet models since those require more GPU RAM. Therefore, we used a server with an Intel i9-9900K CPU (3.6 GHz), 64 GB DDR4 RAM, two RTX 2080ti GPUs with 11.2GB RAM each and an Windows 8.1 64 bit operating system. We implemented the decision tree layer in C++ on the CPU and in CUDA on the GPU. The implementation was integrated into the DLIB~\cite{king2009dlib} framework which uses CUDNN functions. An implementation for Tensorflow~\cite{abadi2016tensorflow} and Torch~\cite{collobert2002torch} is also planned since those are currently the most popular frameworks.

\begin{table}[htb]
	\caption{Comparison between the proposed decision tree layers and convolutions on the MNIST dataset in terms of classification accuracy. As index patterns for the decision tree we used TI2 and TI3. The current sate-of-the-art is shown in gray.}
	\label{tbl:mnist}
	\centering
	\begin{tabular}{lc}
		\toprule
		Method & Result \\
		\midrule
		LeNet-5(TI2) & 99.23 \\
		LeNet-5(TI3) & 99.48 \\
		LeNet-5(Conv.) & 99.37 \\
		\textcolor{gray}{\cite{wan2013regularization}}  & \textcolor{gray}{99.79} \\
		\textcolor{gray}{\cite{cirecsan2012multi}} & \textcolor{gray}{99.77} \\
		\textcolor{gray}{\cite{sato2015apac}} & \textcolor{gray}{99.77} \\
		\bottomrule
	\end{tabular}
\end{table}

Table~\ref{tbl:mnist} shows the results of our adapted Le-Net5 model. As can be seen the TI2 and TI3 pattern (Figure~\ref{fig:arch}) perform similar to the convolutions. The TI2 pattern is an approximation of a $3 \times 3$ convolution and achieves a classification accuracy of 99.23\%. In comparison to this the $5 \times 5$ convolutions as used in the LeNet-5 model achieve an accuracy of 99.37\% which is an improvement of 0.14\%. Approximating the $5 \times 5$ convolutions with the TI3 pattern and the inception technique achieves 99.48. If the runtime is also considered (Table~\ref{tbl:runtime}), it can be seen that the use of the decision trees requires only one third of the computing time in comparison to the convolutions (evaluation on only one CPU core). A disadvantage of the decision trees, on the other hand, is the increased memory consumption. In the case of the LeNet-5 model, both convolution layers require $((1*5*5*6)+6)+((6*5*5*16)+16)=2572$ parameter ($Input * Conv * Output + Bias$). The TI2 pattern needs $((1*1*256*6)+6)+((6*1*256*16)+16)=26134$ parameters and the TI3 pattern needs $((1*6*16*6)+6)+((6*6*16*16)+16)=9814$ ($Input * Inception width * Distribution size * Output + Bias$).

\begin{table}[htb]
	\caption{Comparison between the proposed decision tree layers and convolutions on the CIFAR10 dataset in terms of classification accuracy. As index patterns for the decision tree we used TI1. The current sate-of-the-art is shown in gray.}
	\label{tbl:cifar10}
	\centering
	\begin{tabular}{lc}
		\toprule
		Method & Result \\
		\midrule
		ResNet-34(TI1) & 92.20\\
		ResNet-34(Conv.) & 91.12\\
		\textcolor{gray}{\cite{DBLP:journals/corr/Graham14a}} & \textcolor{gray}{96.53}\\
		\textcolor{gray}{\cite{springenberg2014striving}} & \textcolor{gray}{95.59}\\
		\textcolor{gray}{\cite{mishkin2015all}} & \textcolor{gray}{94.16}\\
		\bottomrule
	\end{tabular}
\end{table}

Table~\ref{tbl:cifar10} shows the comparison between the TI1 pattern and the convolutions using the ResNet-34 model. As can be seen both achieve classification accuracies above 90\%. The TI1 pattern performs slightly better in comparison to the convolutions (1.08 improvement). Comparing the runtime (Table~\ref{tbl:runtime}) of both approaches it can be seen that the decision trees are significantly faster to compute (6.77ms vs 18.1ms). The memory consumption for one distribution of the TI1 pattern is 16 floats, for one convolution 9 floats ($3 \times 3$). This means that the parameters of the model are almost doubled while the runtime is only one third.

\setlength{\tabcolsep}{0.4mm}
\begin{table}[htb]
	\caption{Comparison between the proposed decision tree layers and convolutions on the CIFAR10 dataset in terms of average normalized mean error on the 300W dataset. As index patterns for the decision tree we used TI1. The current sate-of-the-art is shown in gray.}
	\begin{center}
		\begin{tabular}{lccc}
			\toprule
			Method & Comm. & Challe. & Full \\
			\midrule
			ResNet-16(TI1) & 3.64 & 6.81 & 4.26 \\
			ResNet-16(Conv.) & 3.69 & 6.76 & 4.29 \\
			\textcolor{gray}{$SAN_{GT}$~\cite{dong2018style}} & \textcolor{gray}{3.34} & \textcolor{gray}{6.60} & \textcolor{gray}{3.98} \\
			\textcolor{gray}{$SAN_{OD}$~\cite{dong2018style}} & \textcolor{gray}{3.41} & \textcolor{gray}{7.55} & \textcolor{gray}{4.24} \\
			\textcolor{gray}{ResNet-50~\cite{wingloss2018cvpr}} & \textcolor{gray}{3.01} & \textcolor{gray}{6.01} & \textcolor{gray}{3.60}\\
			\textcolor{gray}{CNN-6/7~\cite{wingloss2018cvpr}} & \textcolor{gray}{3.27} & \textcolor{gray}{7.18} & \textcolor{gray}{4.10} \\
			\textcolor{gray}{LAB (8-stack)~\cite{wayne2018lab}} & \textcolor{gray}{3.42} & \textcolor{gray}{6.98} & \textcolor{gray}{4.12} \\
			\textcolor{gray}{3DDFA~\cite{zhu2016face}} & \textcolor{gray}{6.15} & \textcolor{gray}{10.59} & \textcolor{gray}{7.01} \\
			\textcolor{gray}{LBF~\cite{ren2016face}} & \textcolor{gray}{4.95} & \textcolor{gray}{11.98} & \textcolor{gray}{6.32} \\
			\textcolor{gray}{SDM~\cite{xiong2013supervised}} & \textcolor{gray}{5.60} & \textcolor{gray}{15.40} & \textcolor{gray}{7.52} \\
			\bottomrule
		\end{tabular}
	\end{center}
	\label{tbl:cmp300w}
\end{table}

Table~\ref{tbl:cmp300w} shows the results for landmark regression using the ResNet-16 model. As can be seen the convolutions and the decision trees achieve nearly the same result. We used the same pattern (TI1) as for the CIFAR10 classification which means that the memory consumption of the decision trees is nearly twice as high as for the convolutions. The runtime in contrast is halved (4,64ms vs 10.76ms).

\setlength{\tabcolsep}{1.0mm}
\begin{table}[htb]
	\caption{Runtime comparison for different models and index patterns in milliseconds. For the convolutions we used the CUDNN implementation with the best selected algorithm.}
	\label{tbl:runtime}
	\centering
	\begin{tabular}{lccc}
		\toprule
		Model & Input & HW & Runtime\\
		\midrule
		LeNet-5 (TI2) & $28 \times 28 \times 1$ & 1 CPU Core & 0.18ms\\
		LeNet-5 (TI3) & $28 \times 28 \times 1$ & 1 CPU Core & 0.36ms\\
		LeNet-5 (Conv.) & $28 \times 28 \times 1$ & 1 CPU Core & 0.83ms\\
		ResNet-34 (TI1) & $32 \times 32 \times 3$ & GPU 1050ti & 6.77ms\\
		ResNet-34 (Conv.) & $32 \times 32 \times 3$ & GPU 1050ti & 18.1ms\\
		ResNet-16 (TI1) & $80 \times 80 \times 3$ & GPU 1050ti & 4.64ms\\
		ResNet-16 (Conv.) & $80 \times 80 \times 3$ & GPU 1050ti & 10.76ms\\
		\bottomrule
	\end{tabular}
\end{table}

Table~\ref{tbl:runtime} shows an overview of all runtimes of the models used using convolution and decision trees. All runtime evaluations were performed on a single CPU core (Intel i5-4570) or an NVIDIA 1050ti GPU to ensure reproducibility and to simplify the comparison to other hardware environments.

\section{Conclusions and Discussions}
We presented a novel approach for training decision trees in neural network architectures using the back propagation algorithm and showed that it is possible to achieve the same results as with convolutions. Classification and a regression experiment where conducted on publicly available datasets. The improved runtime of the decision trees was estimated theoretically and empirically shown for different models against the high performance CUDNN implementation from NVIDIA. From an industrial point of view, reducing the runtime while maintaining or even improving the predictive quality is a desirable improvement. In contrast to the runtime, the increased memory consumption is a disadvantage. Further research should investigate the use of indexing sets with different depths and the reduction of the decision trees to only necessary paths. Here the authors see further opportunities for the reduction of the computation time and the memory consumption. In addition, the decision trees could also be extended to only use binary weights as it is done in binary convolution neuronal networks~\cite{courbariaux2015binaryconnect}. This would reduce the runtime and memory consumption.

\bibliographystyle{aaai}
\bibliography{AAAI-FuhlW.1964}

\section{Acknowledgments}
Work of the authors is supported by the Institutional Strategy of the University of T\"ubingen (Deutsche Forschungsgemeinschaft, ZUK 63). 

\end{document}